%% file: main.tex
\documentclass[10pt,twocolumn,letterpaper]{article}
\usepackage[accsupp]{axessibility}

\usepackage{cvpr}              
\usepackage[linesnumbered,ruled,vlined]{algorithm2e}
\usepackage{multirow}
\usepackage[most]{tcolorbox}
\usepackage{algorithmicx}
\usepackage{algpseudocode}
\usepackage{tikz}
\usetikzlibrary{calc} 

\definecolor{stepa1}{RGB}{254,230,198}  
\definecolor{stepa2}{RGB}{253,208,160}  

\definecolor{stepb1}{RGB}{232,222,248}  
\definecolor{stepb2}{RGB}{213,199,242}  

\newcommand{\algmark}[1]{\tikz[remember picture,overlay]\coordinate (#1) {};}

\newcommand{\algrect}[4]{%
  \begin{tikzpicture}[remember picture,overlay]
    \path let \p1=(#1), \p2=(#2) in
      coordinate (TL) at ($ (#1) + (0pt, 0.9ex) $)
      coordinate (BR) at ($ (#2) + (\linewidth, \baselineskip - 0.3ex) $)
      coordinate (TR) at ($ (#1) + (\linewidth, 0.9ex) $);
    \fill[#3,opacity=0.42] (TL) rectangle (BR);
    \node[anchor=north east,
          font=\scriptsize,
          fill=white,
          inner xsep=2pt, inner ysep=0.6pt,
          rounded corners=1pt]
      at ($(TR)+(-2pt,0)$) {#4};
  \end{tikzpicture}%
}

\input{preamble}
\definecolor{cvprblue}{rgb}{0.21,0.49,0.74}
\usepackage[pagebackref,breaklinks,colorlinks,allcolors=cvprblue]{hyperref}

\def\paperID{31741} 
\def\confName{CVPR}
\def\confYear{2026}

\title{A Multi-Agent Perception-Action Alliance for Efficient Long Video Reasoning}

\newcommand{\superscr}[1]{\texorpdfstring{$^{#1}$}{(#1)}}
\author{Yichang Xu\superscr{1}, Gaowen Liu\superscr{2}, Ramana Rao Kompella\superscr{2}, Tiansheng Huang\superscr{1}, Sihao Hu\superscr{1} \\
Fatih Ilhan\superscr{1},  Selim Furkan Tekin\superscr{1}, Zachary Yahn\superscr{1}, Ling Liu\superscr{1}\\
\textsuperscript{1}Georgia Institute of Technology, Atlanta, GA\\
\textsuperscript{2}Cisco Systems, USA\\
{\tt\small
xuyichang@gatech.edu, 
\{gaoliu,rkompell\}@cisco.com,} \\
{\tt\small\{sihaohu,thuang,filhan,stekin6,zachary.yahn\}@gatech.edu,
ling.liu@cc.gatech.edu}}

\begin{document}
\maketitle

\input{sec/0_abstract}
\input{sec/1_intro}

\input{sec/method}

\input{sec/exp}
\input{sec/conclusion}

\vspace{12pt}
\noindent {\bf Acknowledgement.\/} 
This research is partially sponsored by NSF CISE grants 2302720 and 2312758, 
a grant from CISCO Edge AI programs, and PACE at the Georgia Institute of Technology. The first and the last author are the primary contacts for this work.


{
    \small
    \bibliographystyle{ieeenat_fullname}
    \bibliography{main}
}

\clearpage
\appendix
\input{sec/appendix}


\end{document}

%% file: sec/0_abstract.tex
\begin{abstract}
\noindent This paper presents a multi-agent perception-action exploration alliance, dubbed A4VL, for efficient long-video reasoning. A4VL operates in a multi-round perception-action exploration loop with a selection of VLM agents. In each round,  the team of agents performs video question-answer (VideoQA) via perception exploration followed by action exploration.  During perception exploration, each agent learns to extract query-specific perception clue(s) from a few sampled frames and performs clue-based alignment to find the video block(s) that are most relevant to the query-specific event. During action exploration, A4VL performs video reasoning in three steps: (1) each agent produces its initial answer with rational, (2) all agents collaboratively scores one another through cross-reviews and relevance ranking, and (3) based on whether a satisfactory consensus is reached, the decision is made either to start a new round of perception-action deliberation by pruning (e.g., filtering out the lowest performing agent) and re-staging (e.g., new-clue and matching block based perception-action exploration), or to conclude by producing its final answer. The integration of the multi-agent alliance through multi-round perception-action exploration, coupled with event-driven partitioning and cue-guided block alignment, enables A4VL to effectively scale to real world long videos while preserving high quality video reasoning. Evaluation Results on five popular VideoQA benchmarks show that A4VL outperforms 18 existing representative VLMs and 11 recent methods optimized for long-video reasoning, while achieving significantly lower inference latency. Our code is released at \url{https://github.com/git-disl/A4VL}.

\end{abstract}

%% file: sec/1_intro.tex
\section{Introduction}
\label{sec:intro}

\begin{figure}
    \centering
    \includegraphics[width=\linewidth]{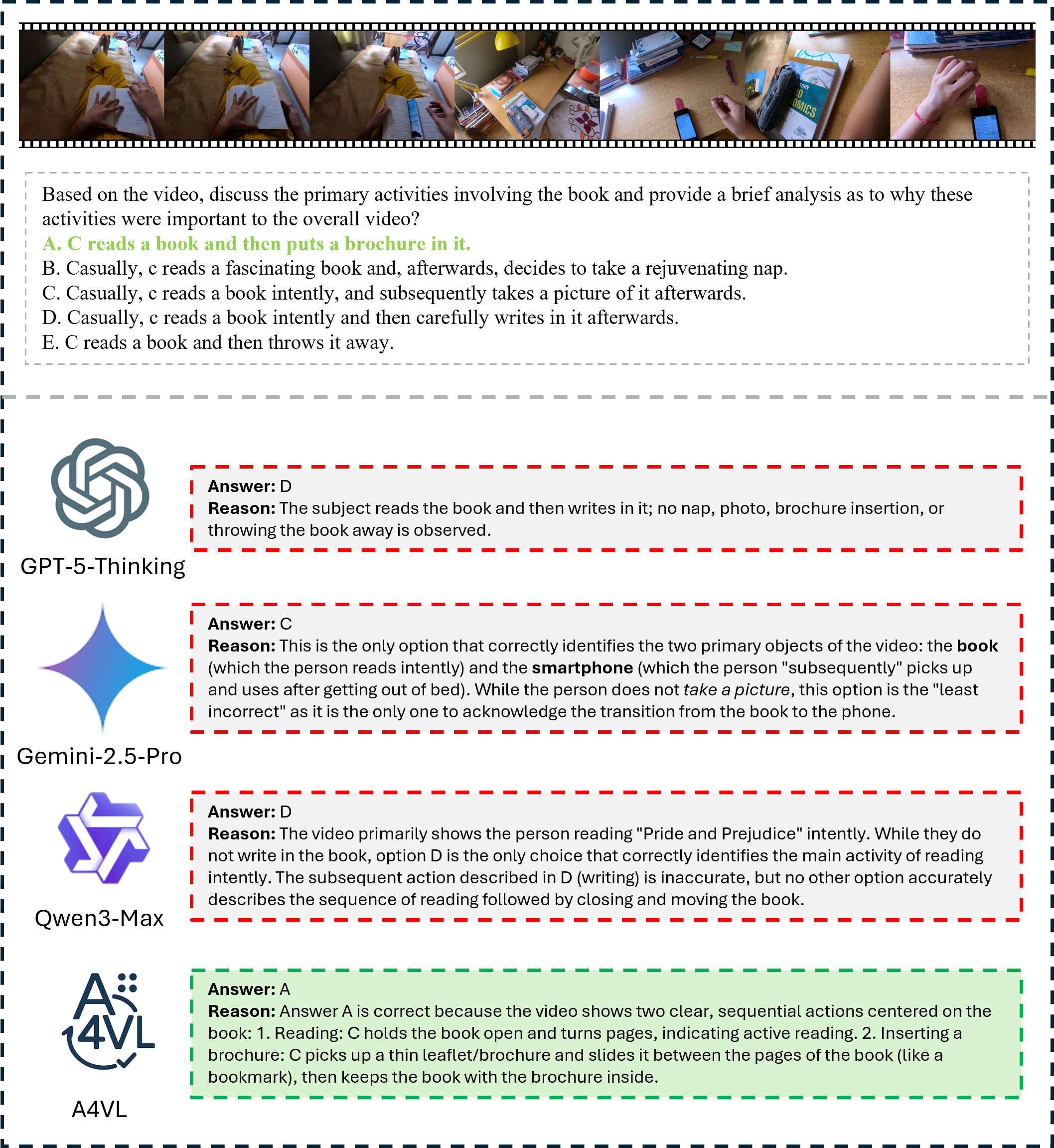}
    \caption{Responses from A4VL and three of the strongest 
    closed-sourced models.}
    \label{fig:example_comparison}
\end{figure}

Multimodal large language models (MLLMs) have been developed to support video understanding in recent years. Compared to traditional approaches~\cite{jang2017tgif, lei2018tvqa, zhu2017uncovering, xue2017unifying, zhao2017video, zhao2017video2}, MLLM-based methods~\cite{team2024gemini, zhang2024videoinstructiontuningsynthetic, damonlpsg2025videollama3, wang2025internvl3_5, zhang2024llavanext-video, qwen2.5-VL, zhang2024long, shen2024longvu, li2024mvbench, li2024llavaonevisioneasyvisualtask, li2024aria} exhibit strong ability for spatiotemporal visual reasoning, as well as human-level text understanding and generation. However, most models often fail to produce high performance reasoning on long videos, because when the number of frames is large, the memory and time costs scale quadratically.
Even with enough computing resources and time, some recent works~\cite{qu2025does, liu2025bolt} show that simply increasing the frame sampling density may even hurt performance, for example, due to redundant frames introducing noise and distracting attention away from truly informative key frames. The problem is especially aggravated for 
video QA workloads that center on questioning events covered in only a small subset of frames, as the problem of identifying accurately where the most relevant frames reside is challenging. Eefficient and scalable solutions are on-demand to improve video understanding performance for complex vision–language reasoning about long and complex videos.

Our work is inspired by some pioneering efforts in this direction. For example, GPT-4o~\cite{openai_gpt4o_system_card_2024} is trained on a large multimodal corpus and, after such training, is able to process long videos by taking many more frames. Although this yields higher accuracy, it is much slower. On Video-MME~\cite{fu2024video}, GPT-4o takes more than 150\,s on average to answer a multiple-choice question. Therefore, the community has begun to explore optimization strategies. Some works focus on token merging~\cite{zhang2025beyond, bao2025dynimg}, sparsification~\cite{korbar2024text, hu2025m, li2024llama, liu2025bolt}, or training-free agent approaches~\cite{suo2025trial, yang2025vca, zhang2025deep, ventura2025chapter, xu2025neurosymbolic,Chen_2025_ICCV,hu2024survey}, while others design more efficient neural architectures~\cite{xu2025auroralong, zhang2025videollama, han2025dynfocus} or memory-retrieval-based systems~\cite{santos2025infty, ma2025drvideo, ye2025re, song2024moviechat}. For example, DYTO~\cite{zhang2025beyond} proposes a dynamic bipartite token merging strategy, which supports a higher frame sampling rate while keeping the resource usage acceptable; AuroraLong~\cite{xu2025auroralong} replaces the LLM component of MLLMs with a linear RNN layer to handle input sequences and also combines token merging techniques to improve efficiency. Among these methods, agent-based approaches and memory retrieval methods are orthogonal to token-based and architecture-focused optimizations. In this paper, we present an innovative  multi-Agent Perception-Action Alliance based method for long video reasoning.

{\bf Problem Statement.\/} We note some limitations of existing agent-based methods. First, their inference speed is low. 
For example, VideoAgent~\cite{fan2024videoagent} takes over ten minutes on a long video that lasts for an hour. Second, most of them use a single MLLM to perform decision making and inference; they do not support effective collaboration among multiple agents. In addition, they often rely on strong video grounding models, and yet when the question is long or complex, they remain struggling to locate key frames. For example, MoReVQA~\cite{min2024morevqa} uses object detection and image QA models to find keyframes and achieves good performance on NeXT-QA~\cite{xiao2021next}. However, its accuracy falls behind other approaches on the long and complex video datasets, e.g.,  EgoSchema~\cite{mangalam2023egoschema} (see Table~\ref{tab:main_results} in Section~\ref{sec:results}). 

To address these limitations, we propose a multi-agent perception–action alliance framework (A4VL), featured with 
three core components: (i) an agent teaming strategy that leverages multiple diverse MLLMs to enhance video reasoning performance; (ii) an event-based partitioning and random sampling strategy to handle resource-intensive long-video processing; and (iii) an iterative agent coordination framework for perception exploration and action exploration that further boosts reasoning performance. 
Figure~\ref{fig:example_comparison} shows the responses from three of the strongest closed-sourced models to date (GPT-5-Thinking~\cite{openai2025gpt5systemcard}, Gemini-2.5-Pro~\cite{comanici2025gemini}, and Qwen3-Max~\cite{qwen3max}), as well as our A4VL, on an example from the EgoSchema dataset. Although all these MLLM-based methods generate incorrect answers, A4VL remains correct. 
Figure~\ref{fig:arch} shows the general framework of A4VL. Figure~\ref{fig:teaming} illustrates our agent teaming strategy, which selects the team that collaborates best from a 
pool of agents. Figure~\ref{fig:example_all_diff} in Section~\ref{sec:method} visualizes the reasoning procedure of A4VL on this example.

Upon receiving a query, A4VL first goes through the perception exploration step. During this step, each agent samples a small number of frames (e.g., 4) to generate a perception clue. We then apply an event-based partition method to divide the video into several blocks and use models like CLIP~\cite{radford2021learning} to obtain the similarity score of each block, measuring its consistency with the perception clue. Each agent samples from high-scoring blocks and forms a set of frames for the next step, e.g., 16 sampled frames. Next, agents move into the action exploration step: they first generate answers and reasons for the sampled frames relevant to a query. If the answers are consistent, a summarizer aggregates the reasons and returns the final answer to the user. Otherwise, each agent is asked to rate every other agent, and the agent team is pruned based on these scores. The surviving agents then revise their perception clues based on the current state and move back to Stage~2 of the perception exploration step, starting the next round of collaboration. 

Our main contributions are summarized as follows:
\begin{itemize}
\item \textbf{A4VL framework.} We develop A4VL, a training\mbox{-}free multi\mbox{-}agent perception–action framework for long-video question answering under tight frame budgets.

\item \textbf{Teaming and clue-guided perception.} We empower A4VL with an unsupervised agent teaming method and an event\mbox{-}based \emph{sample$\rightarrow$clue$\rightarrow$block} pipeline, optimized by combining CLIP-similarity enhanced block selection with adaptive frame allocation, enabling A4VL agents to focus on key frames without retraining.

\item \textbf{Accuracy–latency gains.} On five long-video benchmarks against 28 baselines, A4VL achieves the best accuracy while being substantially faster than strong closed- and open-source MLLMs that process many more frames.
\end{itemize}

%% file: sec/method.tex
\section{Methodology}
\label{sec:method}
\subsection{Design Overview}
\begin{figure*}
    \centering
    \includegraphics[width=\linewidth]{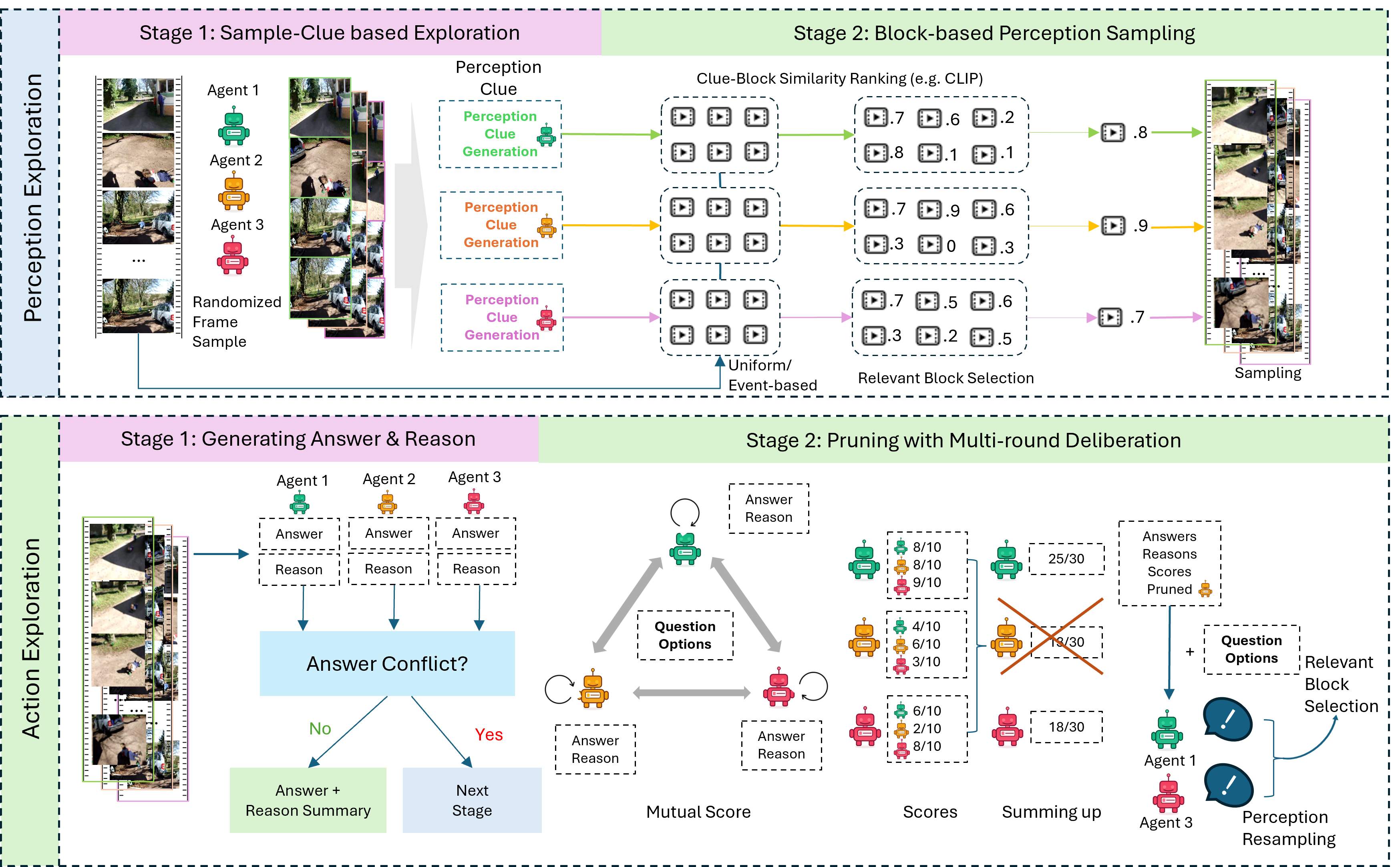}
    \caption{Overview of A4VL architecture.}
    \label{fig:arch}
\end{figure*}

\begin{figure*}[h!]
    \centering
    \includegraphics[width=\linewidth]{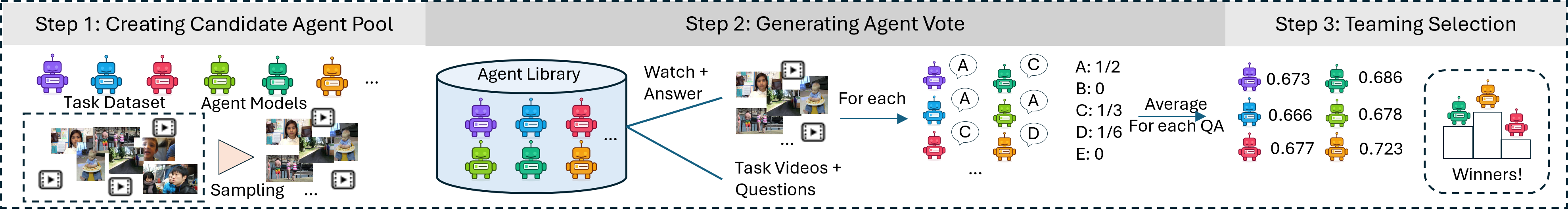}
    \caption{A4VL agent teaming workflow.}
    \label{fig:teaming}
\end{figure*}

\begin{figure*}
    \centering
    \includegraphics[width=\linewidth]{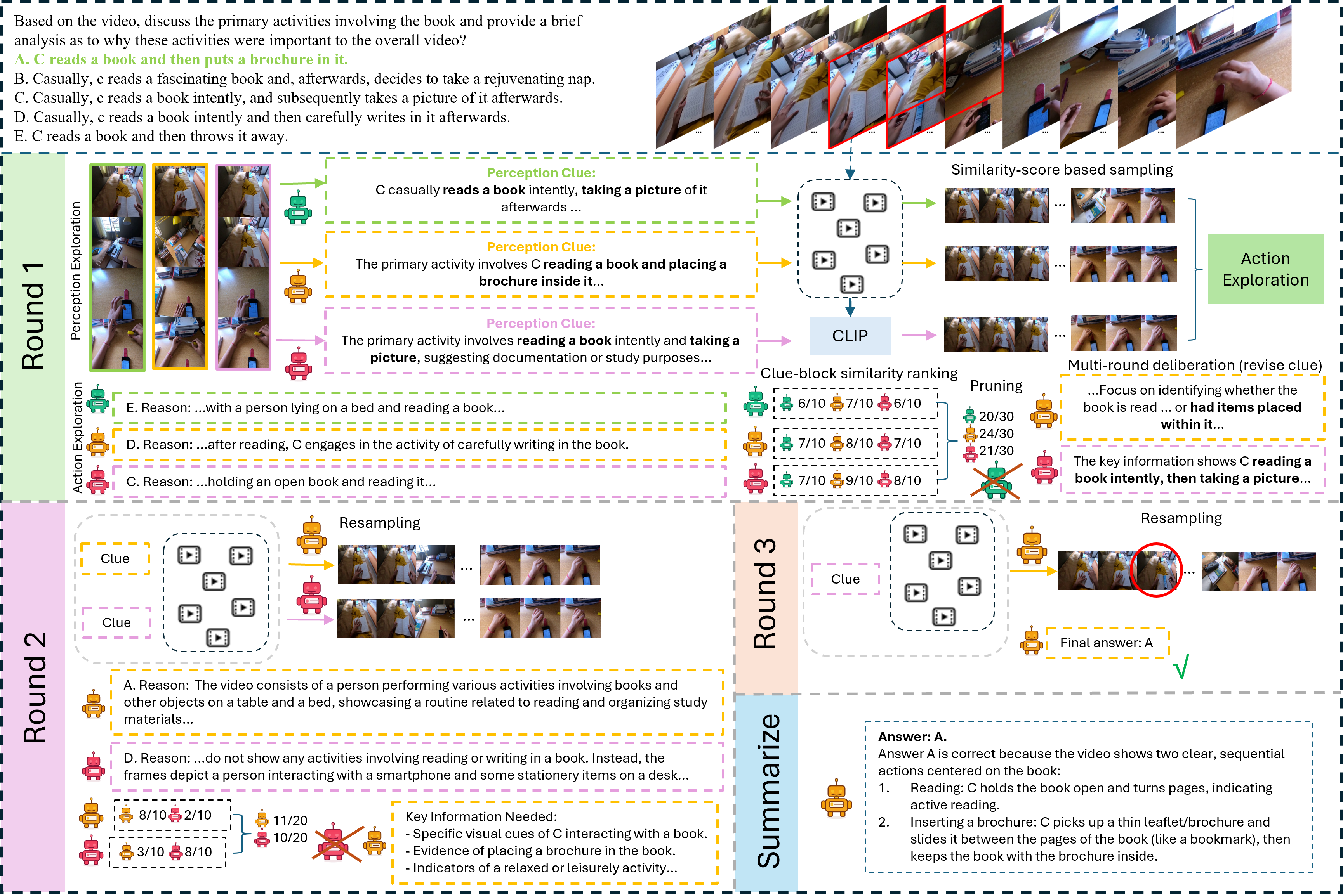}
    \caption{Example of A4VL on EgoSchema. Here all agents have wrong answers at the beginning, but later one of them turns to the correct answer and other agents are pruned.}
    \label{fig:example_all_diff}
\end{figure*}
In this section, we provide an architecture overview of A4VL. As shown in Figure~\ref{fig:arch}, A4VL consists of a perception exploration step and an action exploration step, where multiple agents work collaboratively to produce the final answer. The two steps run iteratively until agents reach a consensus. When we have multiple MLLMs available, we prefer to use a small subset as the agent pool. Hence, we design an optional agent teaming stage (Figure~\ref{fig:teaming}) that selects a task-specific subset of agents. Thereafter, all questions under the same task reuse the same agent subset. The agent teaming process is as follows. Suppose we have $M$ agents and $K$ unlabeled video-question pairs from the task, each with $C$ answer choices. These $K$ samples can be randomly drawn from the task dataset. Using labels would make the teaming algorithm more accurate, but we assume unlabeled data for practical deployment. In the agent teaming step, each agent independently runs the perception exploration and answer generation steps to produce predictions. We count the frequency of each choice in each sample and denote them as $\{f_{qr}\}_{q=1:K,r=1:C}$, where each $f_{qr}$ is the fraction of agents that choose the $r$th option for question $q$. For each agent, let $r_q$ denote the index of the option it selects for question $q$. We define its score as $\frac{1}{K}\sum_{q=1}^Kf_{qr_q}$. We choose $m$ ($m=3$ in our implementation) agents with highest scores to form the task-specific agent pool. As illustrated in Figure~\ref{fig:teaming}, the teaming process has three steps.
\begin{itemize}
    \item \textbf{Step 1}: we prepare the agent library and sample a small set of video-question pairs from the task.
    \item \textbf{Step 2}: each agent runs part of the A4VL workflow and independently generates answers, and we compute the frequency of each answer for each question. For the example in Figure~\ref{fig:teaming}, the frequencies of choices A–E are $1/2$, $0$, $1/3$, $1/6$, and $0$, respectively. Purple, blue, and light green agents therefore receive a score of $0.5$ on that question, while the green and pink agents receive $1/3$, and so on.
    \item \textbf{Step 3}: we average each agent’s score over all sampled questions, and the top 3 agents (green, yellow, and pink) are selected for this task.
\end{itemize}

Within the task, after A4VL receives a video and the user's question (including choices, and additional information like subtitles under some benchmarks), it first goes through the perception exploration step. As shown in Figure~\ref{fig:arch}, the perception exploration step consists of two stages. In Stage~1, each agent randomly samples $N_1$ frames ($N_1=4$ in our settings) to preview the entire video, and then generates a perception clue to identify keyframes. In the illustrative example, the video is divided into six blocks using an event-based approach. For each agent, a CLIP model like~\cite{chen2023attentive} is used to compute similarity scores between its perception clue and each block. These scores indicate how relevant each block is to the current question. Then, $N_2$ frames (set to 16 in our experiments) are sampled from the blocks according to their scores, as described in the perception exploration subsection, and are used in the next action exploration step.

During the action exploration step, each agent is asked to provide an answer and a corresponding reason, given its $N_2$ frames obtained from the previous step. If they reach a consensus, then we give the answer to the user and summarize the reason to explain the answer. Otherwise, each agent is asked to evaluate all agents’ answers, and the agent with the lowest total score (based on self- and peer evaluations) is pruned. For the example shown in Figure~\ref{fig:arch}, the yellow agent is pruned and other agents are required to generate a new perception clue in order to better distinguish between conflicting answers in the next round. After that, all agents will go back to the stage 2 of perception exploration to start a new round, until finally all agents reach a consensus.
\subsection{Perception Exploration}
The top part of Figure~\ref{fig:arch} shows our perception exploration step workflow. We index rounds by $j$. In the first round, each agent goes through the entire perception exploration step; in subsequent rounds ($j>1$), each agent starts directly from selecting relevant blocks. Suppose we have a video $V$ which consists of $n$ frames: $V=\{v_1,v_2,\cdots,v_n\}$, question $Q$, options $O$, and agent pool $\mathbb{A}=\{A_i:1\le i\le m\}$, where $m$ is the number of selected agents after agent teaming. As shown in Figure~\ref{fig:arch}, perception exploration step has two stages: sample-clue based exploration and block-based perception sampling.

In the first stage, each agent generates a perception clue, which contains necessary information that can be used to identify keyframes in stage~2. For each agent $A_i\in\mathbb{A}$, we first sample $N_1$ frames from the video using a sampling strategy $p_1:2^V\times\mathbb{N}^+\rightarrow 2^V$:
\begin{align}
    \hat{v}_{A_i}=p_1(V,N_1).
\end{align}
For $p_1$, we have two choices: (i) we can randomly select $N_1$ frames from the entire video, or (ii) we can divide the video into event-based blocks, and sample a few from each block. In experiments, we test both alternatives and find that (i) is better, so we use choice (i) as our final choice. The reason is that agents do not need fine-grained understanding of the video to generate a coarse perception clue, so uniform random sampling provides higher diversity and better temporal coverage. Given the sampled $N_1$ frames, question, and options, each agent is asked to generate a perception clue, which is later used to identify relevant blocks. We denote the perception clue for agent $A_i$ at $j$th round as $P_{i,j}$. In the first round,
\begin{align}
    P_{i,1}=A_{i,clue}(\hat{v}_{A_i},Q,O).
\end{align}
For subsequent rounds, $P_{i,j}$ is generated at the end of the action exploration step. For example, by the end of the first round, $P_{i,2}$ is generated for each $A_i\in\mathbb{A}$.

In the second stage, we first partition the video into at most $B$ blocks, $B_1,B_2,\cdots,B_b$. We provide two partition strategies: (i) \textbf{uniform partition}, which means $B_i=V_{(i-1)\lfloor\frac{n}{B}\rfloor+1:i\lfloor\frac{n}{B}\rfloor}$; (ii) \textbf{event-based division}, where we detect scene changes using DINOv2 embeddings~\cite{oquab2023dinov2} and simple pixel cues (HSV/motion/sharpness), generate candidates with KTS~\cite{potapov2014category}, PELT~\cite{killick2012optimal}, and SSM-based novelty~\cite{foote2000automatic}, merge them and apply non-maximum suppression (NMS), and keep the top $B-1$ boundaries (at most $B$ blocks). This procedure is efficient, as most videos in our benchmarks can be processed within two seconds. Details are given in the appendix. We also test both strategies in our experiments, and finally choose (ii). After dividing the video into blocks, we compute the similarity between each block and each agent's perception clue. In this paper, we use a CLIP model to compute the similarity scores. Other approaches, such as computing the similarity between the caption of each block and each perception clue, are also possible, but since CLIP is the fastest, we leave the exploration of more advanced similarity computation as future work. We then sample the target frames under two scenarios. (i) If all blocks have similarity scores below $\rho$ (set to 0.8 in our settings), we sample from the most relevant block for each agent: for agent $A_i$, we identify the block $B_{i,j}^\ast$ with the maximum score and use a sampling strategy $p_2:2^V\times\mathbb{N}^+\rightarrow 2^V$ to sample $N_2$ frames from $B_{i,j}^\ast$:
\begin{align}
    V_{act}^{(i)} = p_2(B_{i,j}^\ast, N_2).
\end{align}
(ii) Otherwise, we retain blocks with similarity score $>\rho$. For each agent $A_i$, we obtain a score vector $\mathbf{s}^{(i)}$ over blocks. We first normalize it by $\mathbf{s}^{(i)}\leftarrow\mathbf{s}^{(i)}-\max(\mathbf{s}^{(i)})$, and then compute an integer allocation vector $\mathbf{c}^{(i)}:=\lfloor N_2\cdot SoftMax(\mathbf{s}^{(i)})\rfloor$ for the blocks. If $\sum_k c^{(i)}_k < N_2$, we randomly distribute the remaining $N_2-\sum_k c^{(i)}_k$ samples over the blocks to ensure a total of $N_2$ action frames. We then sample $c^{(i)}_k$ frames from each retained block $B_k$ using $p_2$ and take their union to form $V_{act}^{(i)}$.
\subsection{Action Exploration}
\label{sec:action_expl}
Now each agent has its own set of $N_2$ sampled frames, denoted by $V_{act}^{(i)}$. The action exploration step uses these frames to generate answers and, when necessary, to prune agents. It has two stages: answer and reason generation, and pruning with multi-round deliberation, as shown in the bottom part of Figure~\ref{fig:arch}.

In the first stage, each agent is asked to generate an answer and corresponding reason given its own sampled frames and the question information. For agent $A_i$, it generates the answer $a_{i,j}$ and reason $R_{i,j}$ as follows:
\begin{align}
    a_{i,j}&=A_{i,act}(V_{act}^{(i)},Q,O),\\
    R_{i,j}&=A_{i,reason}(V_{act}^{(i)},a_{i,j},Q).
\end{align}
Then for all agents, we have the answer set $S_{a,j}$ and reason set $S_{r,j}$:
\begin{align}
    S_{a,j}&=\{a_{1,j},a_{2,j},\cdots,a_{|\mathbb{A}|,j}\},\\
    S_{r,j}&=\{R_{1,j},R_{2,j},\cdots,R_{|\mathbb{A}|,j}\}.
\end{align}
If all agents reach a consensus, we send all $P_{i,j}$ values in round $j$, $S_{a,j}$, and $S_{r,j}$ to our reason summarizer to give the user the final answer and an explanation. We consider two ways to check for consensus: (i) \textbf{Majority consensus}: the most frequent answer appears more than $|\mathbb{A}|/2$ times; (ii) \textbf{Full consensus}: all agents output the same answer. We test both strategies and ultimately adopt (ii).

If the agents do not reach a consensus, then we step into the second stage. Given $S_{a,j}, S_{r,j}, Q$ and $O$, each agent evaluates the soundness of its answer and every other agent's answer. It is asked to give a rating from 1 to 10:
\begin{align}
    \{s_{i,\mathbb{A}_1},\cdots,s_{i,\mathbb{A}_{|\mathbb{A}|}}\}\leftarrow A_{i,eval}(Q,S_{a,j},S_{r,j}), \forall A_i\in\mathbb{A}.
\end{align}
In Figure~\ref{fig:arch}, the yellow agent has a rating of 8/10, 6/10 and 2/10 given by three agents. After that, for each agent, we sum up the ratings given by all agents:
\begin{align}
    s_{A_i}:=\sum_{A_k\in\mathbb{A}} s_{k,\mathbb{A}_i}
\end{align}
For example, the yellow agent now has an overall score of 16/30, which is the lowest. Suppose it's $A_2$, then we have $s_{A_2}=16$. Now we do agent pruning, where the agent with the lowest score is removed from the agent pool:
\begin{align}
    \mathbb{A}&\leftarrow \mathbb{A}\setminus\{A_{min}\},\\
    A_{min}&:=\arg\min_{A\in\mathbb{A}}s_A.
\end{align}
In the Figure~\ref{fig:arch} case, $A_{min}=A_2$, and the yellow agent is removed. For each of the remaining agent $A_i$, we generate $P_{i,j+1}$ using $P_{i,j},S_{a,j},S_{r,j},A_{min},Q,O$ to generate the refined perception clue, which is used in the next round:
\begin{align}
    P_{i,j+1}=A_{i,refine}(P_{i,j},S_{a,j},S_{r,j},A_{min},Q,O).
\end{align}

Figure~\ref{fig:example_all_diff} shows an example from EgoSchema. We can see at the beginning, all agents generate different and incorrect answers. However, in the second round, the orange agent begins to realize that choices B, C, D, and E do not have clear visual evidence, and it revises the perception clue to focus on the interaction between the person and the book. In the third round, the keyframe where the person is putting a brochure into the book is sampled (highlighted by the red circle), and therefore the orange agent continues to generate the correct answer. Finally, our summarizer reads the entire multi-round process and returns the final answer together with a concise explanation to the user. The pseudocode of A4VL is provided in the appendix.

%% file: sec/exp.tex
\section{Experiment}
\label{sec:exp}
\begin{table*}[t]
  \centering
  \caption{A4VL compared with other methods on five benchmarks. For Video-MME, the result is in format w. subtitle/wo. subtitle. For Video-MME, the improvement row shows the improvement without subtitle, since A4VL is mainly designed to optimize visual reasoning. * near the model names mean we take the results from the model paper or benchmark leaderboard.}
  {\scriptsize
    \begin{tabular}{ccccccccc}
    \toprule
    \multirow{2}[4]{*}{Model} & \multirow{2}[4]{*}{NeXT-QA} & \multirow{2}[4]{*}{EgoSchema} & \multirow{2}[4]{*}{LongVideoBench} & \multicolumn{1}{c}{\multirow{2}[4]{*}{MLVU}} & \multicolumn{4}{c}{Video-MME} \\
\cmidrule{6-9}          &       &       &       &       & Short & Medium & Long  & Average \\
    \midrule
    GPT-4o*~\cite{openai_gpt4o_system_card_2024} & -     & 72.2  & 66.7  & 54.9  & 80.0/82.8 & 70.3/76.6 & 65.3/72.1 & 71.9/77.2 \\
    Gemini 1.5Pro*~\cite{team2024gemini} & -     & 71.1  & 64.0  &   -    & 81.7/84.5 & 74.3/81.0 & 67.4/77.4 & 75.0/81.3 \\
    \midrule
LLaVA-Video-7B-Qwen2~\cite{zhang2024videoinstructiontuningsynthetic} & 81.2  & 61.8  & 61.1  &  51.7     &  72.4/75.6     &   58.1/63.3    &  50.7/60.4     & 60.4/66.4 \\
    VideoLlama3-7B~\cite{damonlpsg2025videollama3} & 67.4  & 52.6  &   52.5    &  43.9     &  80.1/80.2     &  63.7/69.6     &  54.9/61.0     & 66.2/70.3 \\
    InternVL3.5-8B~\cite{wang2025internvl3_5} & 76.2  & 62.0  &  53.6     &  47.1     &  69.6/74.0 & 56.3/63.9 & 47.7/61.9 & 57.9/66.6 \\
    LLaVA-NeXT-Video-7B~\cite{zhang2024llavanext-video} & 39.7  & 39.6  & 43.5  &  26.3     &  36.4/40.1 & 30.8/38.0 & 27.9/17.2 & 31.7/31.8 \\
    QwenVL-2.5-7B~\cite{qwen2.5-VL} & 76.8  & 60.7  &  54.7     & 47.7      &  72.1/73.6 & 55.7/63.9 & 48.2/60.9 & 58.7/66.1 \\
    LongVA-7B*~\cite{zhang2024long} & 69.3  & -     & 51.3  &   41.1    & 61.1/61.6 & 50.4/53.6 & 46.2/47.6 & 52.6/54.3 \\
    LongVU-7B*~\cite{shen2024longvu} & -     & 37.6  & -     &   -    & -     & -     & -/59.5 & -/60.6 \\
    VideoChat2-7B*~\cite{li2024mvbench} & 61.7  & 56.7  & 39.3  &   30.1   & 48.3/52.8 & 37.0/39.4 & 33.2/39.2 & 39.5/43.8 \\
    \midrule
    LLaVA-Video-72B-Qwen2~\cite{zhang2024videoinstructiontuningsynthetic} & 83.7  & 69.4  &  60.7 &    51.5   &    77.8/80.0 & 64.9/70.8 & 58.6/73.0 & 67.1/74.6 \\
    InternVL3.5-38B~\cite{wang2025internvl3_5} & 78.9  & 75.4  &   57.0    &  56.1     &   76.1/79.1 & 58.6/68.9 & 57.4/68.3 & 64.0/72.1 \\
    InternVL3-78B~\cite{wang2025internvl3_5} & 84.0  & 76.8  &   56.4    &  55.3     &  78.6/80.4     &   65.4/73.4    &   56.7/67.8    & 66.9/73.9 \\
    LLaVA-NeXT-Video-34B~\cite{zhang2024llavanext-video} & 70.1  & 46.6  & 50.5  &  39.5     &   63.1/66.4    &   51.1/53.2    &   44.6/48.7    & 52.5/56.0 \\
    QwenVL-2.5-32B~\cite{qwen2.5-VL} & 76.8  & 64.8  &   53.3    & 48.5      &   73.2/77.1 & 61.6/68.2 & 54.1/68.6 & 63.0/71.3 \\
    QwenVL-2.5-72B~\cite{qwen2.5-VL} & 79.6  & 70.7  &   56.1    & 53.7      &  75.1/77.0 & 66.3/73.3 & 59.6/70.8 & 67.0/73.7 \\
    LLaVA-OneVision-72B*~\cite{li2024llavaonevisioneasyvisualtask} & 80.2  & 62.0  & 63.2  &  -     & 76.7/79.3 & 62.2/66.9 & 66.0/62.4 & 66.3/69.6 \\
    Aria-25.3B*~\cite{li2024aria} & -     & -     & 63.0  &   -    & 67.9/78.3 & 67.0/71.7 & 58.8/66.3 & 67.6/72.1 \\
    \midrule
    ViperGPT~\cite{suris2023vipergpt} & 56.9  & 24.8  &   -    &   24.9    &  -     &   -    &    -   & - \\
    SeViLA~\cite{liu2023deep} & 64.0  & 28.4  &    -   &  31.7     &    -   &    -   &   -    & - \\
    VideoAgent~\cite{fan2024videoagent} & 66.1  & 55.0  &   -    &   34.3    &   -    &   -    &       & - \\
    TraveLER~\cite{shang2024traveler} & 66.0  & 55.4  &   -    &   36.3    &   -    &   -    &    -   &  -\\
    MoReVQA*~\cite{min2024morevqa} & 69.2  & 51.7  & -     & -     & -     & -     & -     & - \\
    VideoRAG-72B*~\cite{luo2024video} & -     & -     & 65.4  &   -    & 81.1/- & 72.9/- & 73.1/- & 75.7/- \\
    AuroraLong*~\cite{xu2025auroralong} & -     & -     & -     & 52.7  & -     & -     & -     & - \\
    DYTO-34B*~\cite{zhang2025beyond} & 72.9  & 56.8  & -     & -     & -     & -     & -     & -/53.4 \\
    BOLT*~\cite{liu2025bolt}  & 79.5  & 64.0  & 59.6  & -     & 70.1/- & 66.0/- & 49.6/- & 59.9/- \\
    DynFocus*~\cite{han2025dynfocus} & -     & -     & 31.8  & 49.6  & 50.9/53.9 & 43.7/46.0 & 37.7/43.6 & 44.1/47.8 \\
    LVAgent~\cite{Chen_2025_ICCV} & 83.0 & 78.4 & 66.9 & 50.0 & 82.4/83.6 & 72.8/78.6 & 66.3/76.4 & 73.9/79.5 \\
    \midrule
    \textbf{A4VL} & \textbf{85.1} & \textbf{82.2} &   \textbf{72.2}    &   \textbf{58.0}    &   \textbf{86.6/87.3}    &  \textbf{76.8/83.2}     &  \textbf{68.3/77.9}     &  \textbf{77.2/82.8} \\
    \textbf{\textcolor{green!70!black}{Improvement$\uparrow$}} & \textbf{\textcolor{green!70!black}{1.9 $\sim$ 45.4}} & \textbf{\textcolor{green!70!black}{3.8 $\sim$ 57.4}} &   \textbf{\textcolor{green!70!black}{5.3 $\sim$ 41.6}}    &   \textbf{\textcolor{green!70!black}{1.9 $\sim$ 19.6}}    &  \textbf{\textcolor{green!70!black}{4.2 $\sim$ 20.2}}     &  \textbf{\textcolor{green!70!black}{2.5 $\sim$ 46.0}}     & \textbf{\textcolor{green!70!black}{0.9 $\sim$ 40.4}} & \textbf{\textcolor{green!70!black}{2.2 $\sim$ 45.5}}  \\
    \bottomrule
    \end{tabular}%
    }
  \label{tab:main_results}%
\end{table*}%
\subsection{Experimental Settings}
We evaluate A4VL on five benchmarks that cover both short and long videos, with an emphasis on long-video reasoning.

To test our method on short videos with strong temporal/causal reasoning requirements, we evaluate A4VL on NeXT-QA~\cite{xiao2021next}. We randomly sample 200 QA pairs per question type from its test split, resulting in a subset of 1{,}493 QA pairs. EgoSchema~\cite{mangalam2023egoschema} is a medium-length video QA dataset consisting of videos lasting about 3 minutes. We evaluate on its test split with ground-truth labels, which contains 500 QA pairs. 

For long videos, we evaluate on LongVideoBench~\cite{wu2024longvideobench}, MLVU~\cite{MLVU} (test split), and Video-MME~\cite{fu2024video}. LongVideoBench consists of 1{,}337 QA pairs, and each video is accompanied by subtitles; many questions are strongly subtitle-related. The videos range from about 3 minutes to 2 hours in length. MLVU-Test is the harder test split of MLVU, containing roughly 500 videos with durations between 3 minutes and 2 hours, and the task is to select the correct answer out of six options. Video-MME consists of 2{,}700 QA pairs and is divided into short, medium, and long groups according to video duration, with 900 QA pairs in each group. Videos in the short group last about 10 seconds, whereas videos in the long group can last for over an hour.

In the agent teaming step, we select $m=3$ agents from a pool of 8 models: LLaVA-Video-7B-Qwen2~\cite{zhang2024videoinstructiontuningsynthetic}, QwenVL-2.5-7B~\cite{qwen2.5-VL}, InternVL3.5-8B~\cite{wang2025internvl3_5}, QwenVL-2.5-32B~\cite{qwen2.5-VL}, InternVL3.5-38B~\cite{wang2025internvl3_5}, InternVL3-78B~\cite{wang2025internvl3_5}, QwenVL-2.5-72B~\cite{qwen2.5-VL}, and LLaVA-Video-72B-Qwen2~\cite{zhang2024videoinstructiontuningsynthetic}. Selecting $m>3$ agents may further boost the accuracy score, but it will harm the efficiency. Unless otherwise specified, all experiments involving A4VL are conducted on six H200 GPUs.

\subsection{Results}
\label{sec:results}
We compare A4VL against two closed-source MLLMs, GPT-4o~\cite{openai_gpt4o_system_card_2024} and Gemini 1.5 Pro~\cite{team2024gemini}, 16 open-source MLLMs, and 10 agent-based or long-video-oriented methods. As shown in Table~\ref{tab:main_results}, A4VL outperforms all baselines on every benchmark. On NeXT-QA, accuracy saturates near 80\% as the backbone size increases; the best baseline, InternVL3-78B, reaches 84.0\%, while A4VL further improves to 85.1\%. On EgoSchema, InternVL3-78B achieves 76.8\%, and A4VL is the only method exceeding 80\%, with 82.2\% (+3.2). LongVideoBench is where we observe the largest gain: compared to GPT-4o at 66.7\%, A4VL (built solely from open-source models) reaches 72.2\% (+5.5). MLVU-Test is substantially harder, with more answer choices and more complex question types (e.g., tracking specific players in long sports videos), yet A4VL still improves over prior methods by 1.9--19.6 points. On Video-MME, A4VL surpasses all baselines across all duration groups and on the overall average. Notably, although the strongest single MLLM backbone reaches at most 67.1\% without subtitles, A4VL attains 77.2\% on average, 2.2 points higher than the best closed-source model, Gemini 1.5 Pro (75.0\%). The models selected by our agent teaming step for each benchmark are listed in Table~\ref{tab:models}.

\begin{table}[htbp]
  \centering
  \caption{Models used for each benchmark. Intern-78B refers to InternVL3-78B~\cite{wang2025internvl3_5}, Intern-38B refers to InternVL3.5-38B~\cite{wang2025internvl3_5}, QwenVL-72B refers to QwenVL-2.5-72B~\cite{qwen2.5-VL}, and LLaVA-72B refers to LLaVA-Video-72B-Qwen2~\cite{zhang2024videoinstructiontuningsynthetic}.}
  {\scriptsize
    \begin{tabular}{c|ccc}
    \toprule
    Benchmark & \multicolumn{3}{c}{Models} \\
    \midrule
    NeXT-QA & Intern-78B & Intern-38B & QwenVL-72B \\
    EgoSchema & Intern-78B & Intern-38B & QwenVL-72B \\
    LongVideoBench & Intern-78B & Intern-38B & QwenVL-72B \\
    MLVU  & Intern-78B & Intern-38B & LLaVA-72B \\
    Video-MME & Intern-78B & LLaVA-72B & QwenVL-72B \\
    \bottomrule
    \end{tabular}%
    }
  \label{tab:models}%
\end{table}%

\begin{table}[htbp]
  \centering
  \caption{Average inference time per sample on four representative methods and A4VL.}
  {\scriptsize
    \begin{tabular}{c|ccc}
    \toprule
    Model & NeXT-QA & EgoSchema & MLVU \\
    \midrule
    GPT-4o & 23s    & 54s    & 127s \\
    InternVL3-78B & 15s    & 50s    & 204s \\
    VideoAgent & 20s    & 83s    & 175s \\
    TraveLER & 101s   & 94s    & 450s \\
    A4VL  & 18s    & 37s    & 74s \\
    \bottomrule
    \end{tabular}%
    }
  \label{tab:efficiency}%
\end{table}%

Table~\ref{tab:efficiency} reports the average inference time per sample for four representative methods---the closed-source MLLM GPT-4o, the strong open-source MLLM InternVL3-78B, and two well-known agent-based systems, VideoAgent and TraveLER---as well as our A4VL. We measure efficiency on NeXT-QA, EgoSchema, and MLVU, which correspond to short-, medium-, and long-video settings. On short videos (NeXT-QA), all methods except TraveLER exhibit similar and acceptable latency. As video length increases, however, the inference time of other methods grows rapidly; for example, on MLVU, even GPT-4o takes over two minutes to answer a single question, whereas A4VL scales much more gracefully. For TraveLER, the inference time on EgoSchema is lower because its official implementation uses image captions only to extract visual information on this dataset, yet A4VL still requires about 60\% less time than TraveLER.

\subsection{Ablation Study}
\begin{figure}
    \centering
    \includegraphics[width=\linewidth]{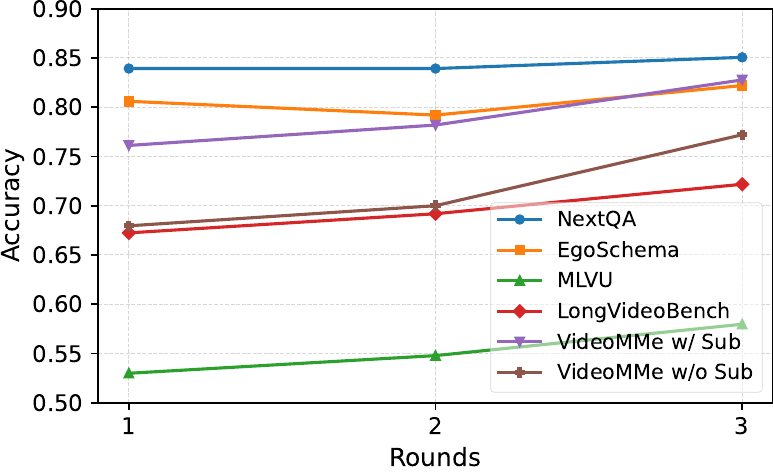}
    \caption{Relationship between accuracy and maximum number of rounds.}
    \label{fig:ablation_round}
\end{figure}

Since multi-round collaboration is a core component of A4VL, we first study how the maximum number of rounds affects accuracy. In our default implementation, A4VL can run at most three rounds because there are three agents in total. In this ablation, we stop the procedure after a fixed maximum round and take the majority answer at that round as the final prediction; if there are multiple majorities, we choose the answer with the highest score. Figure~\ref{fig:ablation_round} shows how accuracy changes as the maximum number of rounds increases. On all benchmarks, accuracy consistently improves when more rounds are allowed. Table~\ref{tab:num_rounds} reports the actual number of rounds used in the default A4VL setting. As the dataset becomes harder, agents tend to collaborate for more rounds. Even within Video-MME, when subtitles are available the task becomes easier and thus requires fewer rounds.

\begin{table}[htbp]
  \centering
  \caption{Number of actual running rounds in A4VL. The data of Video-MME is shown in w/o subtitle / w. subtitle format.}
  {\scriptsize
    \begin{tabular}{c|ccc}
    \toprule
    \#Rounds & 1     & 2     & 3 \\
    \midrule
    NeXT-QA & 1072  & 230   & 191 \\
    EgoSchema & 303   & 101   & 96 \\
    LongVideoBench & 566   & 341   & 430 \\
    MLVU  & 183   & 116   & 203 \\
    Video-MME & 1289/1667 & 557/421 & 854/612 \\
    \bottomrule
    \end{tabular}%
    }
  \label{tab:num_rounds}%
\end{table}%

We also test several design choices discussed in Section~\ref{sec:method}, using EgoSchema as the testbed. We first vary the sampling strategies. For perception-frame sampling $p_1$, we either perform \textbf{r}andom sampling over the entire video (R) or sample one frame from each \textbf{e}vent-based block (E). For action-frame sampling $p_2$, we either sample from uniformly divided blocks (R) or from event-based blocks (E). We evaluate three variants: RRSampling, RESampling, and ERSampling, where the first character denotes $p_1$ and the second denotes $p_2$. A4VL corresponds to RESampling. The accuracy and latency of each variant are shown in Table~\ref{tab:ablation_sampling}. Among all strategies, RESampling achieves the best accuracy with similar latency, so we adopt it as our default. Intuitively, perception exploration benefits from coarse, uniformly distributed frames to cover the whole video, while action exploration benefits from frames concentrated on active events. From the table, we can also know the event-based partitioning only takes about 2 seconds.

\begin{table}[htbp]
  \centering
  \caption{Accuracy and latency of different sampling strategies on EgoSchema. R means random sampling and E means sampling from event-based blocks. The first character represents the sampling strategy for $N_1$ perception frames, and the second character represents the sampling strategy for $N_2$ action frames.}
  {\scriptsize
    \begin{tabular}{c|ccc}
    \toprule
    Method & RRSampling & RESampling & ERSampling \\
    \midrule
    Accuracy (\%) & 80.2  & 82.2  & 79.6 \\
    Time/Sample & 35s    & 37s    & 37s \\
    \bottomrule
    \end{tabular}%
    }
  \label{tab:ablation_sampling}%
\end{table}%

Next, we compare different consensus criteria. Table~\ref{tab:ablation_consens} shows results on EgoSchema under majority consensus (MConsens) and full consensus (FConsens), as defined in Section~\ref{sec:action_expl}. Both standards yield strong accuracy; FConsens has an accuracy of 82.2\%, which is better than MConsensus (81.4\%) but incurs higher (still moderate) latency. The reason is FConsens benefits from increasing collaboration rounds to make sure the final answer is strongly confident. Since the additional latency is acceptable, our default method uses FConsens.

\begin{table}[htbp]
  \centering
  \caption{Results of different consensus conditions on EgoSchema. M means majority and F means full.}
  {\scriptsize
    \begin{tabular}{c|cc}
    \toprule
    Method & MConsens & FConsens \\
    \midrule
    Accuracy (\%) & 81.4  & 82.2 \\
    Time/Sample & 26s    & 37s \\
    \bottomrule
    \end{tabular}%
    }
  \label{tab:ablation_consens}%
\end{table}%

Finally, we test whether agent pruning is necessary. We design two variants without pruning. \textbf{NoPruneSum} selects the answer with the highest total score, and \textbf{NoPruneMaj} selects the majority answer as the final prediction (breaking ties by total score). For both variants, the early stop condition is either (i) all agents in one round generates the same answer (FConsens) or (ii) in two subsequent rounds, the highest score corresponds to the same answer. As shown in Table~\ref{tab:ablation_prune}, no matter how we choose the final answer, removing pruning leads to lower accuracy and substantially higher latency, indicating that pruning is important for both effectiveness and efficiency.

\begin{table}[htbp]
  \centering
  \caption{Results of different pruning strategies on EgoSchema.}
  {\scriptsize
    \begin{tabular}{c|ccc}
    \toprule
    Method & NoPruneSum & NoPruneMaj & A4VL \\
    \midrule
    Accuracy (\%) & 80.8  &   79.4    & 82.2 \\
    Time/Sample & 60s    & 60s    & 37s \\
    \bottomrule
    \end{tabular}%
    }
  \label{tab:ablation_prune}%
\end{table}%

%% file: sec/conclusion.tex
\vspace{-10pt}
\section{Conclusion}
\label{sec:conclusion}

We presented A4VL, a training-free multi-agent framework that tightly couples
perception exploration with action exploration for long video question answering.
By combining event-aware block partitioning, clue-guided frame selection, and
multi-round agent alliance with consensus and pruning, A4VL turns a pool of
heterogeneous MLLMs into a coordinated reasoner over long videos under fixed
compute budgets. Across five different benchmarks, A4VL consistently improves over strong baselines and prior
agent-based systems, and our round-level ablations show stable gains from each collaboration round.

Looking ahead, we believe that perception–action coordination at the agent level
is a promising direction for long-video understanding. Extending A4VL to
audio–text–video tri-modal inputs, richer task-conditioned similarity functions, and neuro-symbolic techniques are natural next steps for enhanced perception exploration toward high performance general long-horizon video agents.

%% file: sec/appendix.tex
\section{A4VL Algorithm}
\label{sec:alg}
The detailed algorithm is shown in Algorithm~\ref{alg:a4vl}. First, the video is partitioned to at most $B$ blocks. Suppose the event-based partitioning algorithm finally divides the video into $b$ ($\le B$) blocks. In case it divides the video into more than $B$ blocks, we simply merge similar blocks until there are $B$ blocks in total. Line 4-9 implements sample-clue based exploration, where the perception clues are generated. Line 10-19 performs block-based perception sampling. Then action exploration step starts from line 20. From line 20-26, each agent generates the answer and the reason, and a consensus check is made to decide whether we early exit. Line 27-34 performs the agent pruning with multi-round deliberation. The worst agent is pruned and remaining agents revise their perception clue for the next round.
\begin{algorithm}[h!]
\caption{A4VL Algorithm.}
\label{alg:a4vl}
\KwIn{Video $V=\{v_1,\dots,v_n\}$, question $Q$, options $O$, agent pool $\mathbb{A}=\{A_i:1\le i\le m\}$, preview frames $N_1$, inference frames $N_2$, max number of blocks $B$.}
\KwOut{Answer $\mathcal{A}$.}

$\{B_1,\dots,B_b\}\leftarrow \text{EventPartition}(V,B)$\\
$j\leftarrow 1$\\
\While{$|\mathbb{A}|\ge 1$}{
  {\algmark{s11}}\ForEach{$A_i\in\mathbb{A}$}{
    \eIf{$j=1$}{
      $\hat{v}_{A_i}\leftarrow p_1(V,N_1)$\\
      $P_{i,1}\leftarrow A_{i,\text{clue}}(\hat{v}_{A_i},Q,O)$
    }{
      $P_{i,j}$ given
    }
  }{\algmark{e11}}
  {\algmark{s12}}\ForEach{$A_i\in\mathbb{A}$}{
    $\mathbf{s}^{(i)}\leftarrow \{Sim(B_k,P_{i,j})|1\le k\le b\}$\\
    \eIf{$\max(\mathbf{s}^{(i)})>\rho$}{$\mathbf{s}^{(i)}[\mathbf{s}^{(i)}\le\rho]=-\infty$\\
    $\mathbf{s}^{(i)}\leftarrow\mathbf{s}^{(i)}-\max(\mathbf{s}^{(i)})$\\
    $\mathbf{c}^{(i)}\leftarrow N_2\lfloor SoftMax(\mathbf{s}^{(i)})\rfloor$\\
    $V^{(i)}_{\text{act}}\leftarrow\bigcup_{k=1}^b p_2(B_k,c^{(i)}_k)$}{$k^\ast\leftarrow \arg\max_{k\in\{1,\dots,b\}} s^{(i)}_k)$\\
    $V^{(i)}_{\text{act}}\leftarrow p_2(B_{k^\ast},N_2)$}
  }{\algmark{e12}}
  {\algmark{s21}}\ForEach{$A_i\in\mathbb{A}$}{
    $a_{i,j}\leftarrow A_{i,\text{act}}(V^{(i)}_{\text{act}},Q,O)$\\
    $R_{i,j}\leftarrow A_{i,\text{reason}}(V^{(i)}_{\text{act}},a_{i,j},Q)$
  }
  $S_{a,j}\leftarrow \{a_{i,j}\mid A_i\in\mathbb{A}\}$\\
  $S_{r,j}\leftarrow \{R_{i,j}\mid A_i\in\mathbb{A}\}$\\
  \If{$\forall i,i':\, a_{i,j}=a_{i',j}$}{
    \Return{$\mathcal{A}\leftarrow a_{1,j}$}
  }{\algmark{e21}}
  {\algmark{s22}}\ForEach{$A_i\in\mathbb{A}$}{
    $\left(s_{i,\mathbb{A}_1},\dots,s_{i,\mathbb{A}_{|\mathbb{A}|}}\right)\leftarrow A_{i,\text{eval}}(Q,S_{a,j},S_{r,j})$
  }
  \ForEach{$A_i\in\mathbb{A}$}{
    $s_{A_i}\leftarrow \sum_{A_k\in\mathbb{A}} s_{k,\mathbb{A}_i}$
  }
  $A_{\min}\leftarrow \arg\min_{A\in\mathbb{A}} s_A$\\
  $\mathbb{A}\leftarrow \mathbb{A}\setminus\{A_{\min}\}$\\
  \ForEach{$A_i\in\mathbb{A}$}{
    $P_{i,j+1}\leftarrow A_{i,\text{refine}}(P_{i,j},S_{a,j},S_{r,j},A_{\min},Q,O)$
  }{\algmark{e22}
  $j\leftarrow j+1$}
  }
\Return{$\mathcal{A}\leftarrow a_{i^\star,j-1}$} \tcp{$A_{i^\star}$ is the last remaining agent}
\end{algorithm}

\algrect{s11}{e11}{stepa1}{Step 1.1: Sample-Clue Based Exploration}
\algrect{s12}{e12}{stepa2}{Step 1.2: Block-Based Perception Sampling}

\algrect{s21}{e21}{stepb1}{Step 2.1: Generating Answer \& Reason}
\algrect{s22}{e22}{stepb2}{Step 2.2: Pruning with Deliberation}

\section{Additional Visualizations}
\begin{figure}[h!]
    \centering
    \includegraphics[width=\linewidth]{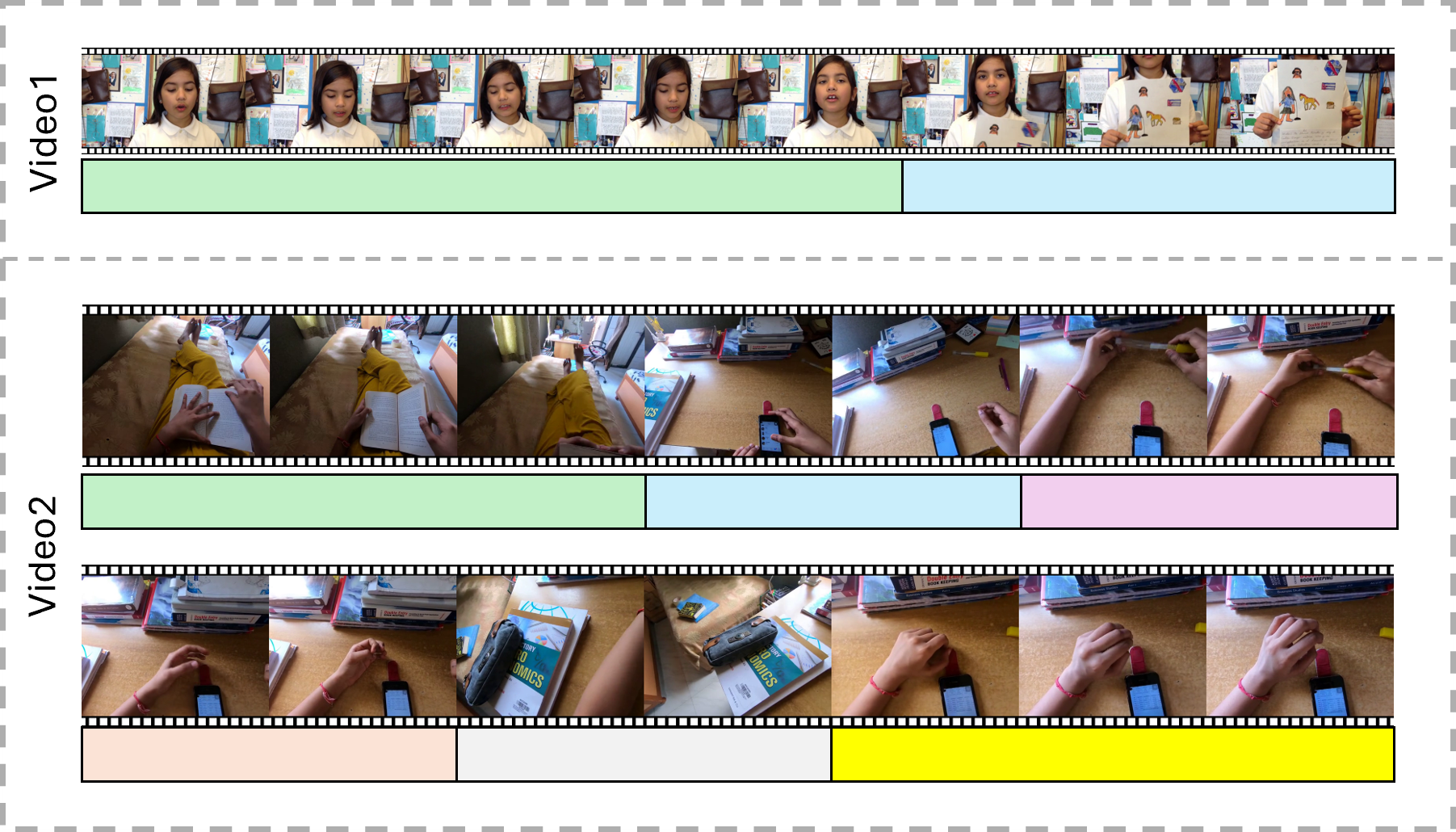}
    \caption{Visualization of event-based video partitioning.}
    \label{fig:partition}
\end{figure}

Figure~\ref{fig:partition} shows the result of event-based partitioning on two cases. Segments are annotated by colored blocks below the video frames. The top case is a video from NeXT-QA, which lasts for 25 seconds. Our event-based partitioning algorithm accurately divides the video into two blocks: in the first block, the girl keeps talking, and in the second block, the girl takes out a paper. The bottom example is from EgoSchema, which lasts for three minutes. Our partitioning algorithm divides the video into six blocks, where the person alternates between reading the book, using the phone, taking out a pen, and looking around. The entire procedure is training free and query agnostic.
\begin{figure*}[h!]
    \centering
    \includegraphics[width=\linewidth]{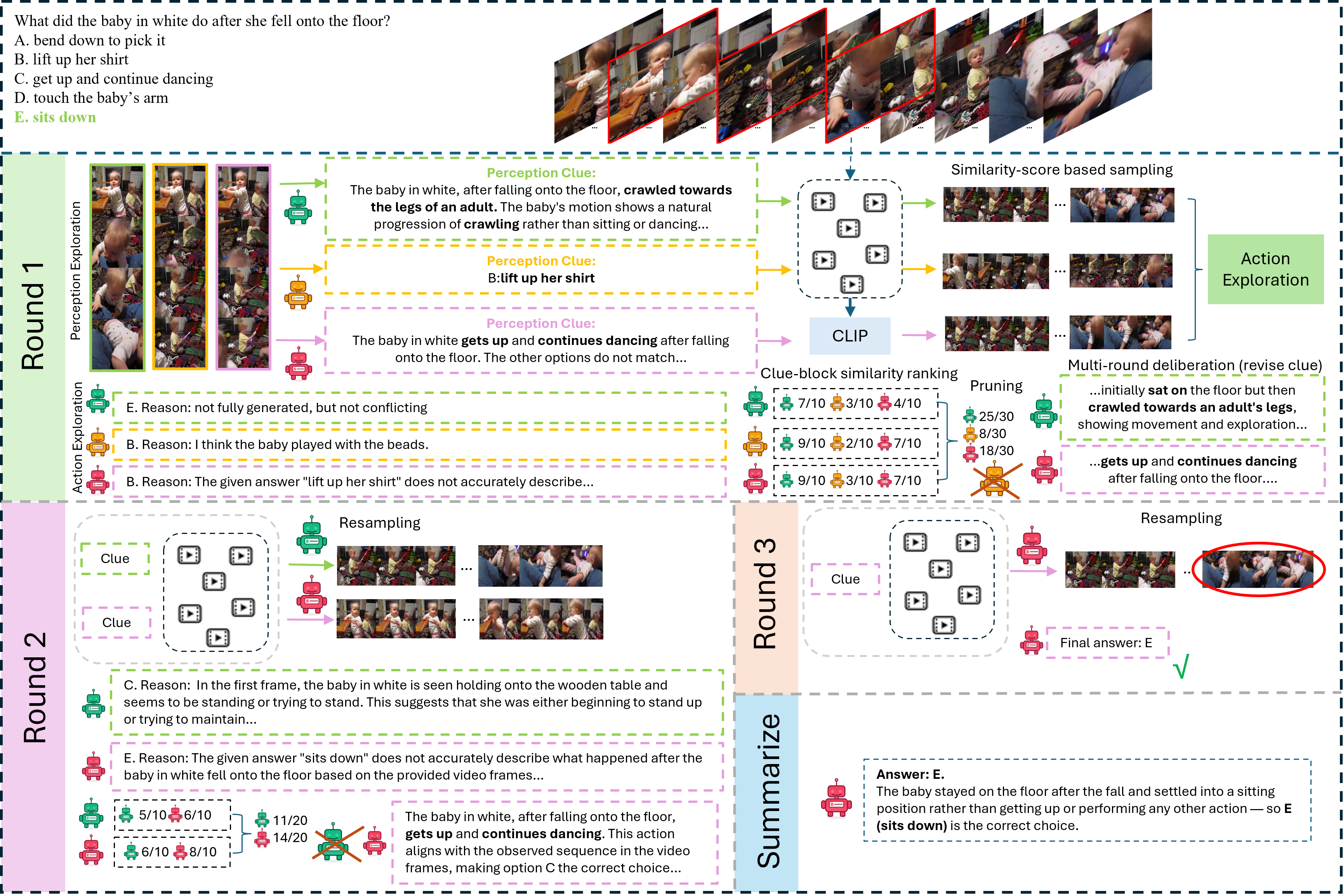}
    \caption{Example where the majority is wrong in the first round, but finally A4VL produces the correct answer. The example is taken from NeXT-QA.}
    \label{fig:example_maj_wr}
\end{figure*}
Figure~\ref{fig:example_maj_wr} presents a case in which the majority answer (B) in the first round is incorrect, yet A4VL ultimately returns the correct answer in the final round, supported by a sound explanation from our summarizer. In this example, the green agent corresponds to InternVL3.5-38B~\cite{wang2025internvl3_5}, the orange agent to InternVL3-78B~\cite{wang2025internvl3_5}, and the pink agent to QwenVL-2.5-72B~\cite{qwen2.5-VL}. We observe that, in the first round, only the green agent produces the correct answer, while the other two receive significantly lower scores. Notably, the green agent is not the model with the largest number of parameters. However, it exhibits instability and becomes incorrect in the second round, resulting in its removal. In contrast, the pink agent becomes correct in the second round and remains correct through the third round, leading to the correct final answer. These visualizations complement Figure~\ref{fig:arch} to~\ref{fig:example_all_diff} in the main paper by illustrating how event-based partitioning and multi-round collaboration behave on real videos.

\section{Failure Cases}
\begin{figure*}[h!]
    \centering
    \includegraphics[width=\linewidth]{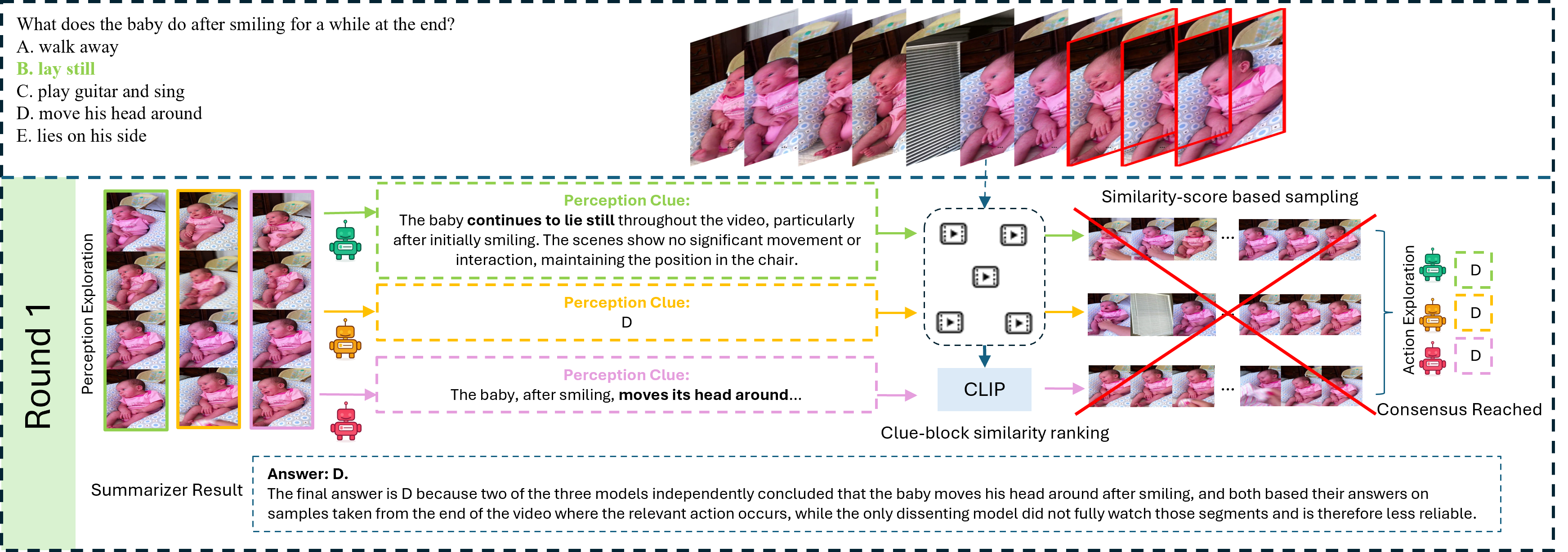}
    \caption{Failure case where all agents achieve a wrong consensus in the first round. The example is taken from NeXT-QA.}
    \label{fig:fail_all_wr}
\end{figure*}
Although A4VL demonstrates superior performance over existing models, it can still fail in certain cases. For example, when all agents make an incorrect prediction in the first round, the early-exit mechanism forces the final answer to be wrong. Figure~\ref{fig:fail_all_wr} shows such a failure case: the final answer is incorrectly chosen as option~D because all agents in the first round output~D. This error arises from several factors: (i) it is difficult for the perception frames to accurately capture the moments after the baby smiles; (ii) although the green agent generates the perception clue ``lay still,'' CLIP is an image--text similarity model and thus struggles to capture temporal movements such as ``still,'' leading to the selection of an incorrect block, while the other two agents, whose perception clues are themselves incorrect, also choose wrong blocks; and (iii) the question references the video position ``at the end,'' but CLIP cannot measure the temporal alignment between frames and text. Addressing such cases is a challenging but important direction for future work, and potential remedies include integrating more advanced video grounding models and routing complex cases to these models when detected, as well as incorporating neuro-symbolic techniques for precise video processing, such as explicitly cropping the end segment of the video.

\begin{figure*}[h!]
    \centering
    \includegraphics[width=\linewidth]{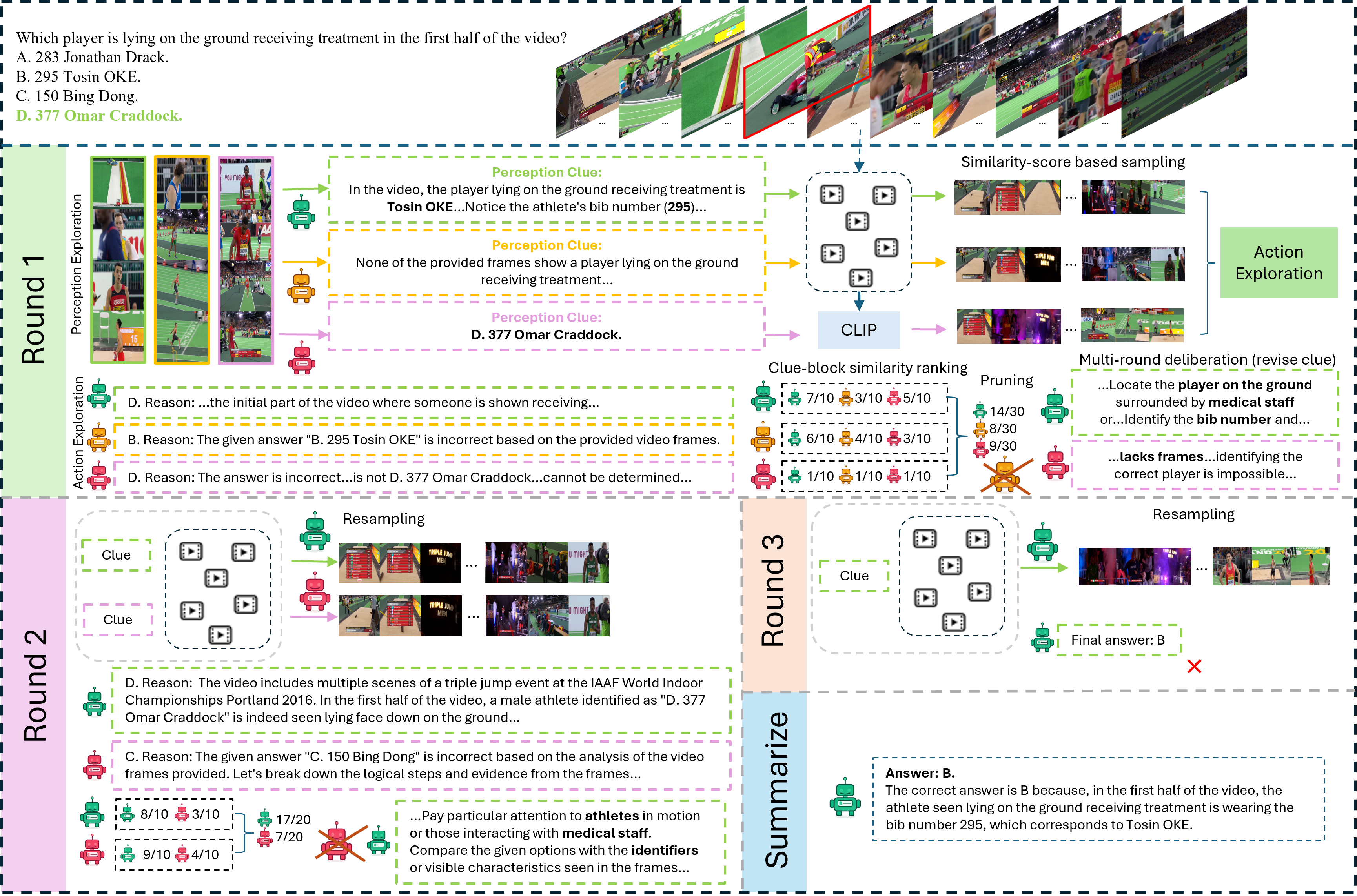}
    \caption{Failure case where the majority is correct in the first round, but finally the final answer gets wrong. This example is taken from Video-MME (w/o sub).}
    \label{fig:fail_maj_wr}
\end{figure*}
It is also possible that the majority vote in the first round is correct, while the final answer is wrong. As shown in Figure~\ref{fig:fail_maj_wr}, we illustrate such a case from Video-MME (without subtitle). The green, orange, and pink agents represent InternVL3-78B~\cite{wang2025internvl3_5}, QwenVL-2.5-72B~\cite{qwen2.5-VL}, and LLaVA-Video-72B-Qwen2~\cite{zhang2024videoinstructiontuningsynthetic}, respectively. This is a challenging example because the entire video is about 20 minutes long, while the keyframes last only a few seconds. The perception frames provide no useful clues, and none of the sampled action frames contain the keyframe where a player is lying on the ground receiving treatment. As a result, each agent’s behavior is highly unstable when reasoning over these irrelevant frames. In the first round, two agents choose answer D, likely because that player happens to appear most frequently and is often centered in the view. However, their predictions shift in rounds 2 and 3, eventually converging on the wrong answer B. Another difficulty is that the question explicitly restricts the focus to the first half of the video. If this constraint were reliably detected and only the first half were used in the multi-round reasoning process (e.g., via the neural-symbolic approach mentioned earlier), the outcome might improve, though sampling the true keyframes would remain non-trivial. For such questions, an efficient framework based on fine-grained watching is therefore needed.

\section{Event-Based Partitioning}
\label{sec:appendix_partition}

Given a video $V$, our goal is to obtain a small set of temporal boundaries that partition $V$ into at most $B$ visually coherent blocks $\{B_1,\dots,B_b\}$, where $b \le B$. We follow a training-free, query-agnostic pipeline: (i) sampled decoding and feature extraction, (ii) construction of a fused novelty signal on time, (iii) multi-scale pooling and PELT-based segmentation on the 1D signal, and (iv) embedding-based heads (SSM and KTS) whose outputs are combined with non-maximum suppression (NMS) and a cap of at most $B-1$ cuts.

\vspace{0.3em}
\paragraph{Sampled frames and notation.}
Let $V$ have $T$ frames $\{I_1,\dots,I_T\}$ with timestamps $0 \le t_1 < \dots < t_T = L$. We uniformly sample at most $F$ frames (in practice $F=200$) and denote their indices by
\[
\mathcal{S} = \{i_1 < i_2 < \dots < i_N\}, \quad N \le F,
\]
with corresponding images $I_{i_r}$ and timestamps $t_{i_r}$. For readability, we abbreviate $I_r := I_{i_r}$ and $t_r := t_{i_r}$ when referring to the sampled frames.

\vspace{0.3em}
\paragraph{Color, sharpness, motion, and embeddings.}
For each sampled frame $I_r$, we compute four types of features.

\textbf{HSV color histograms.}
we detect color changes across frames using histogram distances in HSV space. Specifically, we convert $I_r$ from BGR to HSV, obtaining $H_r^{\text{img}}(x)$, $S_r^{\text{img}}(x)$, and $V_r^{\text{img}}(x)$ for pixel locations $x \in \Omega$ (image grid). We discretize each channel into $K$ bins (we use $K=32$) and compute histograms
\[
h_r^H \in \mathbb{R}^K,\quad
h_r^S \in \mathbb{R}^K,\quad
h_r^V \in \mathbb{R}^K,
\]
followed by a joint $\ell_1$ normalization:
\[
\tilde{h}_r^H = \frac{h_r^H}{S_r},\quad
\tilde{h}_r^S = \frac{h_r^S}{S_r},\quad
\tilde{h}_r^V = \frac{h_r^V}{S_r},
\]
with
\[
S_r = \sum_{k=1}^K h_r^H[k] + \sum_{k=1}^K h_r^S[k] + \sum_{k=1}^K h_r^V[k] + \varepsilon,
\]
where $\varepsilon>0$ is a small constant for numerical stability. The concatenated color descriptor is
\[
c_r = \big[\tilde{h}_r^H;\,\tilde{h}_r^S;\,\tilde{h}_r^V\big] \in \mathbb{R}^{3K}.
\]

\textbf{Sharpness.}
we detect focus changes using the variance of a Laplacian filter. We convert $I_r$ to gray-scale $G_r(x)$ and apply a discrete Laplacian filter $\Delta G_r(x)$ (e.g., the standard $3\times 3$ kernel). The sharpness score is the variance of the Laplacian:
\[
\mu_r^{\Delta} = \frac{1}{|\Omega|} \sum_{x \in \Omega} \Delta G_r(x), \qquad
s_r = \frac{1}{|\Omega|} \sum_{x \in \Omega} \big(\Delta G_r(x) - \mu_r^{\Delta}\big)^2.
\]

\textbf{Motion.}
we measure frame-to-frame intensity differences as a proxy for motion magnitude. For $r \ge 2$, we define a robust frame-to-frame motion magnitude using the median absolute difference between consecutive gray-scale images:
\[
d_r^{\text{mot}} = \operatorname{median}_{x \in \Omega} \big| G_r(x) - G_{r-1}(x) \big|, \qquad
m_r = \frac{1}{255}\, d_r^{\text{mot}},
\]
and set $m_1 = 0$.

\textbf{DINOv2 embeddings.}
we use cosine distance in feature space to capture semantic changes: we pass a resized RGB version of $I_r$ through a DINOv2 encoder~\cite{oquab2023dinov2} and obtain a feature vector
\[
e_r \in \mathbb{R}^D.
\]
We use $\ell_2$-normalized embeddings $\hat{e}_r = e_r / (\lVert e_r \rVert_2 + \varepsilon)$ when computing cosine similarities.

\vspace{0.3em}
\paragraph{Per-step novelty from consecutive frames.}
We now define novelty cues on the temporal midpoints
\[
\tau_r = \frac{t_{r-1} + t_r}{2}, \qquad r=2,\dots,N.
\]

\textbf{Color novelty.}
For each channel we use cosine distance between histograms; for hue, for example:
\[
d_r^H = 1 - \cos\!\big(\tilde{h}_r^H, \tilde{h}_{r-1}^H\big)
      = 1 - \frac{\langle \tilde{h}_r^H, \tilde{h}_{r-1}^H \rangle}
                {\lVert \tilde{h}_r^H \rVert_2 \,\lVert \tilde{h}_{r-1}^H \rVert_2 + \varepsilon},
\]
and analogously $d_r^S$ and $d_r^V$. We combine them as
\[
c_r^{\text{nov}} = \alpha\, d_r^H + \beta\, d_r^S + \gamma\, d_r^V.
\]
where $\alpha=0.55, \beta=0.35$ and $\gamma=0.10$ in our settings.

\textbf{Sharpness and motion novelty.}
We use absolute temporal differences:
\[
u_r^{\text{shp}} = \big| s_r - s_{r-1} \big|, \qquad
u_r^{\text{mot}} = \big| m_r - m_{r-1} \big|.
\]

\textbf{Embedding novelty with EMA.}
We first compute a forward exponential moving average (EMA) of embeddings:
\[
\tilde{e}_1 = e_1, \qquad
\tilde{e}_r = \delta \tilde{e}_{r-1} + (1-\delta) e_r,
\]
with $\delta = 0.9$. For each $r \ge 2$ we define two cosine distances:
\[
d_r^{\text{prev}} = 1 - \cos(\hat{e}_r, \hat{e}_{r-1}), \qquad
d_r^{\text{ema}}  = 1 - \cos(\hat{e}_r, \widehat{\tilde{e}}_{r-1}),
\]
where $\widehat{\tilde{e}}_{r-1}$ is the $\ell_2$-normalized EMA. The embedding novelty is
\[
u_r^{\text{emb}} = \tfrac{1}{2} d_r^{\text{prev}} + \tfrac{1}{2} d_r^{\text{ema}}.
\]

\vspace{0.3em}
\paragraph{Robust normalization and fusion.}
For any scalar sequence $x_r$ ($r=2,\dots,N$) we denote its \emph{robust $z$-score} by
\[
\operatorname{med}(x) = \operatorname{median}_{r} x_r,\]
\[
\operatorname{MAD}(x) = \operatorname{median}_{r} \big| x_r - \operatorname{med}(x) \big|,
\]
\[
z_r(x) = \frac{x_r - \operatorname{med}(x)}{1.4826\,\operatorname{MAD}(x) + \varepsilon}.
\]
The constant $1.4826 = 1/\Phi^{-1}(0.75)$, where $\Phi$ is the standard normal CDF, so that
$1.4826\,\mathrm{MAD}$ is a consistent estimator of the standard deviation for Gaussian data.

We additionally apply a short moving average of window $w$ (we use $w=5$) to reduce noise:
\[
\bar{z}_r(x) = \frac{1}{|\mathcal{W}_r|} \sum_{u \in \mathcal{W}_r} z_u(x),
\]
where $\mathcal{W}_r = \{u : |u-r|\le \lfloor w/2 \rfloor\}$ truncated at the sequence boundaries. We finally clip values to $[-4,4]$.

We apply this procedure separately to the color, motion, sharpness, and embedding novelties:
\[
z_r^{\text{col}} = \bar{z}_r\big(c^{\text{nov}}\big), \quad
z_r^{\text{mot}} = \bar{z}_r\big(u^{\text{mot}}\big),\]
\[
z_r^{\text{shp}} = \bar{z}_r\big(u^{\text{shp}}\big), \quad
z_r^{\text{emb}} = \bar{z}_r\big(u^{\text{emb}}\big).
\]

We then compute cue-specific empirical standard deviations
\[
\sigma_{\text{col}} = \operatorname{std}_r\big(z_r^{\text{col}}\big), \quad
\sigma_{\text{mot}} = \operatorname{std}_r\big(z_r^{\text{mot}}\big)\]
\[
\sigma_{\text{emb}} = \operatorname{std}_r\big(z_r^{\text{emb}}\big), \quad
\sigma_{\text{shp}} = \operatorname{std}_r\big(z_r^{\text{shp}}\big)
\]
and combine them with fixed base weights
\[
\beta_{\text{col}} = 0.20,\quad
\beta_{\text{mot}} = 0.30,\quad
\beta_{\text{emb}} = 0.35,\quad
\beta_{\text{shp}} = 0.05.
\]
The final fusion weights are
\[
w_c = \frac{\beta_c \sigma_c}{\sum_{c' \in \{\text{col},\text{mot},\text{emb},\text{shp}\}} \beta_{c'} \sigma_{c'}}.
\]

The fused novelty signal on midpoints $\tau_r$ is
\[
s(\tau_r) = 
w_{\text{col}}\, z_r^{\text{col}} +
w_{\text{mot}}\, z_r^{\text{mot}} +
w_{\text{emb}}\, z_r^{\text{emb}} +
w_{\text{shp}}\, z_r^{\text{shp}}.
\]
We apply another short moving average in time (again with window $5$) to obtain a smooth 1D signal, which we still denote by $s(\tau_r)$.

\vspace{0.3em}
\paragraph{Multi-scale pooling and PELT heads.}
To handle varying frame rates and video lengths, we pool $s(\tau_r)$ onto several temporal grids. For a grid resolution $\Delta > 0$ (we use $\Delta \in \{0.25, 0.5, 1.0\}$ seconds), we define bin edges
\[
u_0 = \tau_2,\quad u_J = \tau_N,\quad
u_j = u_0 + j\Delta,
\]
and bins $[u_j, u_{j+1})$ for $j = 0,\dots,J-1$. The pooled signal is
\[
s_j^{(\Delta)} = \max\big\{ s(\tau_r) \;:\; \tau_r \in [u_j,u_{j+1}) \big\},
\]
with bin centers $t_j^{(\Delta)} = (u_j + u_{j+1})/2$.

On each pooled sequence $\{s_j^{(\Delta)}\}_{j=0}^{J-1}$ we apply PELT~\cite{killick2012optimal} with a standard $\ell_2$ cost and a penalty $\lambda$ swept over a small range. Denoting by $\mathcal{K}^{(\Delta)}$ the predicted change-point indices for a given $\lambda$, the breakpoints correspond to bin centers $\{t_k^{(\Delta)} : k \in \mathcal{K}^{(\Delta)}\}$. We only keep candidates that yield segments of length at least
\[
\ell_{\min} = \max\left(\ell_0,\; \frac{L}{15}\right),
\]
where $\ell_0$ is a user-defined minimum (e.g., $2$\,s). For each candidate index $k$ we estimate a simple boundary strength via the difference of local averages:
\[
\gamma_k^{(\Delta)} = \left| \frac{1}{W} \sum_{j=k}^{k+W-1} s_j^{(\Delta)} - \frac{1}{W} \sum_{j=k-W}^{k-1} s_j^{(\Delta)} \right|,
\]
with a small window $W$ (e.g., $W \approx \ell_{\min}/\Delta$). We discard boundaries with $\gamma_k^{(\Delta)}$ below a quantile threshold and collect the remaining breakpoints from all grid levels into a set $\mathcal{T}_{\text{PELT}}$.

\vspace{0.3em}
\paragraph{Embedding-based SSM and KTS heads.}
We additionally operate directly on the embedding sequence. We first resample embeddings $\{e_r\}$ to a grid $\{\tilde{t}_q\}$ at roughly $1$\,Hz and obtain averaged embeddings $\tilde{e}_q$ in each bin.

\textbf{SSM-based novelty.}
We $L_2$-normalize $\tilde{e}_q$ to $\hat{e}_q$ and build the Gram matrix
\[
S_{pq} = \langle \hat{e}_p, \hat{e}_q \rangle.
\]
We then follow Foote's self-similarity matrix (SSM) novelty detector~\cite{foote2000automatic}. For each center index $c$ we extract a $(2w)\times(2w)$ block around $(c,c)$, partition it into four $w\times w$ quadrants, and apply a Gaussian-weighted checkerboard kernel $K$ with $+1$ on the top-left and bottom-right quadrants and $-1$ on the others:
\[
n_c = \sum_{i,j} K_{ij}\, S_{(c+i-w),(c+j-w)}.
\]
We robustly normalize the sequence $\{n_c\}$ via the same MAD-based $z$-score and keep local maxima separated by at least $\ell_{\min}$ and above a quantile threshold. The corresponding times form $\mathcal{T}_{\text{SSM}}$. We set $\ell_{\min}=\max\{l,4\}$, where $l$ is the video length.

\textbf{KTS-style segmentation.}
We also run a KTS-like greedy procedure~\cite{potapov2014category}. Let $E$ be the matrix whose rows are $\hat{e}_q$. We form the Gram matrix $G = E E^\top$ and its integral image to enable $O(1)$ evaluation of segment similarity. For a segment $[a,b]$, we define the cost
\[
C(a,b) = -\frac{1}{(b-a+1)^2} \sum_{p=a}^b \sum_{q=a}^b G_{pq}.
\]
Starting from the full interval $[0,Q-1]$, we recursively search for a split index $k$ that maximizes the gain
\[
\text{gain}(a,b,k) = C(a,b) - \big( C(a,k) + C(k+1,b) \big) - \lambda_{\text{KTS}},
\]
with a data-dependent penalty $\lambda_{\text{KTS}}$ (set to 1.4 in our experiment, under all benchmarks). If the best gain is positive and both resulting segments respect $\ell_{\min}$, we split and recurse. The resulting cut times form $\mathcal{T}_{\text{KTS}}$.

\vspace{0.3em}
\paragraph{Merging, NMS, and truncation.}
We merge all candidate boundaries
\[
\mathcal{T}_{\text{all}} = \mathcal{T}_{\text{PELT}} \cup \mathcal{T}_{\text{SSM}} \cup \mathcal{T}_{\text{KTS}},
\]
sort them increasingly, and apply a one-dimensional non-maximum suppression with minimum separation $\ell_{\min}$. Concretely, we scan $\mathcal{T}_{\text{all}}$ in order and keep a candidate $\tau$ if $\tau - \tau_{\text{last}} \ge \ell_{\min}$, where $\tau_{\text{last}}$ is the time of the last kept boundary; otherwise we discard it. This yields a filtered set $\tilde{\mathcal{T}}$ and ensures that every resulting segment has duration at least $\ell_{\min}$.

If $|\tilde{\mathcal{T}}| > B-1$, we compute a boundary strength measure (e.g., $\gamma_k^{(\Delta)}$ on a fixed grid) for each $\tau \in \tilde{\mathcal{T}}$, rank boundaries by strength, and keep only the top $B-1$. Adding the implicit boundaries at $0$ and $L$, we obtain
\[
0 = \tau_0 < \tau_1 < \dots < \tau_{b-1} < \tau_b = L,
\]
which define the final blocks $B_k = V[\tau_{k-1},\tau_k)$ used by A4VL.

The overall algorithm is shown in Algorithm~\ref{alg:event}.

\begin{algorithm}
\caption{Event-based video partitioning (\textsc{EventPartition}).}
\label{alg:event}
\KwIn{Video $V$, max blocks $B$, max feature frames $F$, min segment length $\ell_{\min}$.}
\KwOut{Blocks $\{B_1,\dots,B_b\}$ with $b \le B$.}

\tcp{Step 1: sampled decode \& feature extraction}
Uniformly sample at most $F$ frames from $V$ and record $\{I_r,t_r\}_{r=1}^N$\;
\For{$r=1$ \KwTo $N$}{
  Compute $c_r,G_r,s_r,m_r,e_r,\hat{e}_r$ as in Appendix.~\ref{sec:appendix_partition}\;
}

\tcp{Step 2: per-step cues and fused novelty}
\For{$r=2$ \KwTo $N$}{
  Compute color novelty $c_r^{\text{nov}}$, $u_r^{\text{shp}} = |s_r-s_{r-1}|$, $u_r^{\text{mot}} = |m_r-m_{r-1}|$, and $u_r^{\text{emb}}$ from embeddings (previous \& EMA)\;
  Set midpoint $\tau_r = (t_{r-1}+t_r)/2$\;
}
Apply MAD-based robust $z$-scoring and short moving average to obtain $z_r^{\text{col}},z_r^{\text{mot}},z_r^{\text{shp}},z_r^{\text{emb}}$\;
Compute fusion weights $w_c$ and fused novelty $s(\tau_r) = \sum_c w_c z_r^{(c)}$, then smooth in time\;

\tcp{Step 3: PELT heads on pooled grids}
$\mathcal{T}_{\text{PELT}} \leftarrow \emptyset$\;
\ForEach{$\Delta \in \{0.25,0.5,1.0\}$}{
  Pool $s(\tau_r)$ into bins of width $\Delta$ (max pooling) to get $s_j^{(\Delta)}$\;
  Run PELT on $\{s_j^{(\Delta)}\}_j$ with segment length $\ge \ell_{\min}$ and keep strong boundaries (via local strength $\gamma_k^{(\Delta)}$)\;
  Add resulting times to $\mathcal{T}_{\text{PELT}}$\;
}

\tcp{Step 4: embedding-based SSM and KTS heads}
Resample $\{e_r\}$ to a $\sim 1$\,Hz grid $\{\tilde{e}_q,\tilde{t}_q\}$\;
Compute SSM-based novelty on the Gram matrix of $\{\tilde{e}_q\}$ and keep strong peaks (gap $\ge \ell_{\min}$) to form $\mathcal{T}_{\text{SSM}}$\;
Run KTS-style greedy segmentation on $\{\tilde{e}_q\}$ with penalty $\lambda_{\text{KTS}}$ to obtain $\mathcal{T}_{\text{KTS}}$\;

\tcp{Step 5: merge, NMS, and truncation}
$\mathcal{T}_{\text{all}} \leftarrow \mathcal{T}_{\text{PELT}} \cup \mathcal{T}_{\text{SSM}} \cup \mathcal{T}_{\text{KTS}}$\;
Sort $\mathcal{T}_{\text{all}}$ and apply 1D NMS with minimum gap $\ell_{\min}$ to obtain $\tilde{\mathcal{T}}$\;
\If{$|\tilde{\mathcal{T}}| > B-1$}{
  Rank boundaries by strength and keep the top $B-1$\;
}
Let $\tau_0 = 0$, $\tau_b = L$, and $\{\tau_1,\dots,\tau_{b-1}\} = \tilde{\mathcal{T}}$ in order\;
\Return{$B_k = V[\tau_{k-1},\tau_k)$, $k=1,\dots,b$}\;
\end{algorithm}
\section{Related Works}
\paragraph{Video understanding.} After the development of image QA, researchers began to explore video QA as a more complex multimodal understanding task. Pioneering works~\cite{jang2017tgif, lei2018tvqa, zhu2017uncovering, xue2017unifying, zhao2017video, zhao2017video2} mainly adopt simple CNN–RNN style pipelines that encode pre-extracted frame features with LSTMs/GRUs over short clips, offering only limited modeling of long-range temporal structure and multimodal context. Modern methods begin to use LLM backbones for more powerful multimodal understanding. Representative multimodal LLMs such as GPT-4o~\cite{openai_gpt4o_system_card_2024}, Gemini~\cite{team2024gemini,comanici2025gemini}, LLaVA-OneVision / LLaVA-NeXT-Video~\cite{li2024llavaonevisioneasyvisualtask,zhang2024llavanext-video}, InternVL3.5~\cite{wang2025internvl3_5}, Qwen2.5-VL / Qwen3-Max~\cite{qwen2.5-VL,qwen3max}, VideoLLaMA3~\cite{zhang2025videollama}, and ARIA~\cite{li2024aria} couple strong language backbones with unified visual encoders, long-context modeling, and instruction tuning, substantially advancing open-ended image and long-form video understanding. Also, agent-based methods like VideoAgent~\cite{fan2024videoagent}, TraveLER~\cite{shang2024traveler} or MoReVQA~\cite{min2024morevqa} use multi-step reasoning for tasks with strong temporal or causal relationship.

\paragraph{Long video understanding.}
Recent benchmarks such as EgoSchema~\cite{mangalam2023egoschema}, LongVideoBench~\cite{wu2024longvideobench}, MLVU~\cite{MLVU}, and Video-MME~\cite{fu2024video} push models to reason over minutes- to hour-long videos with complex queries and interleaved modalities, exposing the limitations of naive uniform sampling and single-pass decoding~\cite{mangalam2023egoschema, li2024mvbench, MLVU, fu2024video}. To improve scalability, a line of work compresses or sparsifies the visual stream via token merging, adaptive compression, and content-aware resampling~\cite{bao2025dynimg, hu2025m, korbar2024text, li2024llama, li2024llama, li2024aria, zhang2024long, zhang2025beyond}, while others design specialized architectures or training objectives, e.g., RNN-based MLLMs and self-reward–aligned long-video learners~\cite{shen2024longvu, suo2025trial, xu2025auroralong}. Retrieval- and memory-based methods instead treat long videos as external knowledge to be indexed, using document retrieval, video-RAG, or sparse memory to focus reasoning on a small subset of salient clips~\cite{wu2024longvideobench, ma2025drvideo, luo2024video, song2024moviechat, zhang2025deep}. Closest to ours are agentic or search-style systems that iteratively query long videos, such as VCA, MovieChat, Chapter-LLaMA, $\infty$-Video, and related multi-agent frameworks~\cite{song2024moviechat, yang2025vca, ventura2025chapter, santos2025infty, fan2024videoagent, shang2024traveler, min2024morevqa}. In contrast, A4VL is a training-free multi-agent alliance that combines event-based partitioning with clue-guided block selection and multi-round agent collaboration, and can be plugged on top of arbitrary MLLM backbones to improve long-video reasoning under strict frame budgets.

\paragraph{Agents in video understanding.}
Agent-based frameworks have recently been introduced for video understanding, where multiple specialized modules (``agents'') collaborate via planning, tool use, and iterative querying. For short videos, systems such as ViperGPT~\cite{suris2023vipergpt}, MoREVQA~\cite{min2024morevqa}, and Traveler~\cite{shang2024traveler} decompose questions into sub-tasks and route them to dedicated reasoning or perception components, often improving compositionality and interpretability over clip-level inputs. For long video understanding, BOLT~\cite{liu2025bolt}, VCA~\cite{yang2025vca}, VideoAgent~\cite{fan2024videoagent}, Deep Video Discovery~\cite{zhang2025deep}, MovieChat~\cite{song2024moviechat}, and related agentic pipelines employ global controllers, external memory, and iterative search over the temporal axis to locate relevant segments under strict token budgets. Our A4VL follows this agentic perspective but remains training-free, combining event-based partitioning, clue-guided block selection, and multi-agent consensus to robustly discover key evidence in very long videos.